\documentclass[a4paper,twoside]{article}

\usepackage{epsfig}
\usepackage{subcaption}
\usepackage{calc}
\usepackage{amssymb}
\usepackage{amstext}
\usepackage{amsmath}
\usepackage{amsthm}
\usepackage{multicol}
\usepackage{pslatex}
\usepackage{apalike}
\usepackage{algorithm2e}
\usepackage[bottom]{footmisc}
\usepackage{graphicx}
\usepackage{multirow}
\usepackage{array}
\usepackage{url}
\usepackage[table]{xcolor}% http://ctan.org/pkg/xcolor
\usepackage{color, colortbl}
\usepackage{comment}

\newcommand{\degree}{^{\circ}}

\DeclareMathOperator*{\argmax}{arg\,max}

\usepackage{SCITEPRESS}     % Please add other packages that you may need BEFORE the SCITEPRESS.sty package.

\begin{document}

% Create a framed box with the copyright text
\noindent
\begin{minipage}{\textwidth}
\begin{center}
\fbox{%
  \parbox{0.85\textwidth}{%
    \textbf{Copyright Notice} \\
    \copyright\ 2024 International Conference on Informatics in Control, Automation and Robotics (ICINCO)\\
    This is the author’s version of an article that has been accepted to ICINCO 2024 (SciTePress proceedings). Changes will be made to this version by the publisher prior to publication.\\
    The proceedings version will be distributed as open access under CC BY-NC-ND 4.0 license. No further permission to reuse this article is required. For other type of use, contact SciTePress, Lda.
  }
}
\end{center}
\end{minipage}

\title{Model-Free versus Model-Based Reinforcement Learning for Fixed-Wing UAV Attitude Control Under Varying Wind Conditions}

\author{\authorname{David Olivares\sup{1}\sup{2}\orcidAuthor{0009-0001-5021-3042}, Pierre Fournier\sup{1}\orcidAuthor{0000-0002-0209-2901}, Pavan Vasishta\sup{2}\orcidAuthor{0000-0002-2976-8457} and Julien Marzat\sup{1}\orcidAuthor{0000-0002-5041-272X}}
\affiliation{\sup{1}DTIS, ONERA, Université Paris-Saclay, 91123 Palaiseau, France}
\affiliation{\sup{2}AKKODIS Research, 78280 Guyancourt, France}
\email{\{david.olivares, pierre.fournier, julien.marzat\}@onera.fr, pavan.vasishta@akkodis.com}
}


\abstract{This paper evaluates and compares the performance of model-free and model-based reinforcement learning for the attitude control of fixed-wing unmanned aerial vehicles using PID as a reference point. The comparison focuses on their ability to handle varying flight dynamics and wind disturbances in a simulated environment. Our results show that the Temporal Difference Model Predictive Control agent outperforms both the PID controller and other model-free reinforcement learning methods in terms of tracking accuracy and robustness over different reference difficulties, particularly in 
nonlinear flight regimes. Furthermore, we introduce actuation fluctuation as a key metric to assess energy efficiency and actuator wear, and we test two different approaches from the literature: action variation penalty and conditioning for action policy smoothness. We also evaluate all control methods when subject to stochastic turbulence and gusts separately, so as to measure their effects on tracking performance, observe their limitations and outline their implications on the Markov decision process formalism.}

\onecolumn \maketitle \normalsize \setcounter{footnote}{0} \vfill

\section{\uppercase{Introduction}}
\label{sec:introduction}

Robotic controllers must act according to their state and environment. In the context of fixed-wing unmanned aerial vehicles (FWUAV) flight, modeling the aircraft and its interactions with the environment is a challenging task for two reasons. 

First, the aerodynamic forces exhibit different dynamic regimes that depend on the attitude of the FWUAV. On one hand, low attitude angles near the equilibrium state present relatively simple dynamics often approximated by formulating hypotheses on their linearity and independent coupling of the controlled axes. On the other hand, high attitude angles lead the FWUAV to encounter complex aerodynamics where lift and drag behave nonlinearly and also exhibit important cross-couplings between axes.

A second source of complexity lies in FWUAVs potentially evolving in disturbed environments. These disturbances are the result of unpredictable meteorological phenomena such as turbulence and wind gusts. As a result, effectively modeling FWUAVs' aerodynamics is often a tedious task required to design Control Theory (CT)-based controllers.

A solution could come from Reinforcement Learning (RL) to alleviate the burden of system modeling by using data to approximate these models either implicitly (model-free RL) or explicitly (model-based RL). 

% \begin{figure}[h]
%     \centering
%     \includegraphics[width=0.40\textwidth]{img/x8_insim.png}
%     \caption{The Skywalker x8 FWUAV in JSBSim simulation}
%     \label{fig:x8_insim}
% \end{figure}

Model-free RL (MF-RL) control for attitude control of a quadrotor UAV is presented as outperforming the widely used PID in \cite{koch2019reinforcement}. Similar conclusions are drawn by \cite{song2023reaching} for acrobatic high-speed UAV racing. For FWUAVs, MF-RL was introduced as a robust control approach to deal with turbulence \cite{bohn2019deep}. As for model-based RL (MB-RL), it showed recent breakthroughs by outperforming MF-RL algorithms in highly complex dynamical environments such as games, locomotion and manipulation tasks \cite{hafner2023dreamerv3,hansen2024tdmpc2}. We identify a gap in literature for MB-RL applied to UAV flight: most recent works focus on MB-RL's sample efficiency for learning flying policies \cite{becker2020learning,lambert2019low} and do not evaluate the performance of latest MB-RL algorithms. While not being the focus of our study, MB-RL often incorporates concepts from Control Theory (CT) which can mitigate the blackbox aspect of MF-RL (by adding a planning step) and could be an additional benefit of this class of methods. Our work aims to evaluate the benefits of RL methods, in particular MB-RL versus MF-RL, for attitude control of a FWUAV under varying specific conditions (wind disturbances).

We focus on the attitude control problem, where the controller has to track reference attitude angles for roll and pitch. First, we choose state-of-the-art MF-RL methods such as Proximal Policy Optimization (PPO) and Soft Actor Critic (SAC). Second, we select Temporal Difference Model Predictive Control (TD-MPC), a MB-RL method that mixes concepts from RL such as temporal difference learning and concepts from CT for predictive planning. Finally, we include a PID controller as a reference point, because of its very wide use in UAV attitude control.

Our contributions are as follows: 
\begin{itemize}
    \item We propose the first application of a recent MB-RL method (TD-MPC) to FWUAV attitude control under varying wind conditions.
    \item Our comparison provides insights for FWUAV practitioners on the current state-of-the-art RL algorithms with respect to the industry standard PID.
    \item We exhibit TD-MPC's superiority for pure reference tracking in the case of nominal wind conditions (no turbulence and no gusts), most notably on hard attitude angle references.
    \item We improve this analysis with two additional studies:
    \begin{itemize}
        \item A secondary metric for actuation fluctuation shows that RL, be it model-free or model-based, struggles to produce smooth action outputs. We evaluate two methods from the literature to deal with this issue.
        \item In the presence of wind perturbations and when evaluated specifically on this aspect, the gains of RL methods appear more limited than suggested by previous work \cite{bohn2019deep}.
    \end{itemize}
    \item For replicability and facilitating comparison among contributors, we release our code for the RL-compatible simulation framework\footnote{\texttt{https://github.com/Akkodis/FW-JSBGym}} based on the open-source flight simulator JSBSim \cite{berndt2004jsbsim} and the control algorithms\footnote{\texttt{https://github.com/Akkodis/FW-FlightControl}} presented in this work.
\end{itemize}

\section{RELATED WORKS}
\label{sec:related-works}

% \subsection{Classical Nonlinear Control of UAVs}
% To account for the nonlinearities of a fixed-wing UAV, nonlinear control methods like feedback linearization \cite{kawakami2017nonlinear}, sliding mode control \cite{castaneda2017extended} or linear parameter varying (LPV) control~\cite{rotondo2017lpv} have been applied. Despite the benefits nonlinear control brings, it strongly relies on accurate modeling of the entire system, which can be a challenging task and can result in a controller too sensitive to model uncertainty. In addition, UAV control systems need to operate in real time with limited computational resources. Consequently, optimization-based controllers that need resource consuming sophisticated models or online optimization such as Nonlinear Model Predictive Control (NMPC) \cite{kang2009linear} have limited applications.
Learning-based methods appear as a promising tool to achieve nonlinear and perturbation resilient controllers for UAVs. An example of learning with CT is the field of adaptive control, which now includes learning-based approaches where a neural network learns a model of the problematic part of the dynamics (disturbances, modeling errors) and uses it as a feed-forward component inside a classical control controller \cite{shi2019neural,doukhi2019neural,shi2021omac}. These studies displayed encouraging results for jointly using a learned model of the unknown dynamics together with a mathematical model of the known nominal dynamics.

\subsection{Reinforcement Learning Control of UAVs}
However, when the nominal dynamics are not given, RL has proven to be a powerful choice for learning controllers without any prior knowledge of the system. It has shown impressive results for decision making in games \cite{mnih2013atari,schrittwieser2020mastering} and in robotics \cite{hansen2024tdmpc2,hafner2023dreamerv3} such as dexterous robotic hand manipulation  \cite{andrychowicz2020learning} or quadruped locomotion~\cite{lee2020learning,peng2020learning} and can be classified into model-free and model-based approaches.
Since rotary-wing and fixed-wing UAVs share similarities, methods applicable to one may transfer effectively to the other.

\subsubsection{Model-Free RL}
In model-free RL (MF-RL), the agent learns to take actions through trial and error without learning or using an explicit model of the environment's dynamics. For rotary-wing UAVs, MF-RL has been applied as a flight controller for waypoint navigation \cite{hwangbo2017control} and the recent work of \cite{koch2019reinforcement} has applied various RL algorithms in order to learn attitude control policies outperforming their baseline PID controller counterpart. A proof-of-concept Deep Deterministic Policy Gradient (DDPG) agent has demonstrated encouraging attitude control performance despite exhibiting steady state errors \cite{tsourdos2019developing}. Recent work has highlighted the benefits of RL for trajectory tracking over classical two-stage controllers (a high-level stage for trajectory planning and a lower-level stage for attitude control) \cite{song2023reaching}. The authors concluded that the cascading of two control loops splits the overall waypoint tracking objective into two sub-objectives leading to suboptimal solutions, whereas RL can optimize the higher-level waypoint objective directly.

For FWUAVs, experiments have been conducted in the context of high performance control in a combat environment of an F-16 fighter jet \cite{de2023deep}. RL has also been studied for addressing specific problems on civil aircraft \cite{liu2021learning}, where the authors used vision-based observations to achieve takeoff under crosswind conditions. A study \cite{bohn2019deep} compared RL and PID attitude controllers under wind disturbances and both methods achieved similar performance, paving the way for the deployment of a RL control law in a real FWUAV~\cite{bohn2023data}. The former \cite{bohn2019deep} demonstrated PPO's resilience to stochastic turbulence in simulation and the latter demonstrated SAC as a data-efficient algorithm for real-world deployment of the RL attitude controller. We use those two works as starting points and extend the analysis with a MB-RL method (TD-MPC) and under an additional disturbance source: wind gusts.

\subsubsection{Model-Based RL}
 Model-based RL (MB-RL) is a branch of RL where the agent, through trial and error too, separately learns a model of the environment and an action planning strategy from this model. MB-RL is often more sample efficient, requiring fewer training episodes than MF-RL algorithms to converge. This observation was exploited in \cite{becker2020learning} where the authors trained a controller for waypoint navigation of a rotary-wing UAV, directly in the real world. \cite{lambert2019low} present a sample efficient MB-RL method for hovering control using a ``random-shooter" MPC for simulating candidate trajectories obtained with a learned-model of the dynamics. A higher-level trajectory tracking off-line MB-RL controller is presented by \cite{liang2018scalablembrl}, leveraging data trajectories obtained on multiple UAVs. Current MB-RL literature for UAVs emphasizes MB-RL's sample efficiency for learning flying policies, yet does not assess the performance of the latest MB-RL algorithms. Moreover, we found no studies regarding MB-RL applied to FWUAV flight under varying wind conditions. With the present paper, we intend to extend the existing literature by evaluating the benefits of MF-RL versus MB-RL for this specific setting.
 
%===============================================================================

\section{UAV MODEL}
\label{sec:uav-model}
\subsection{Kinematics}
The studied FWUAV is the Skywalker x8 as in \cite{bohn2019deep}. The FWUAV is modeled as a rigid-body mass $m$ with inertia matrix~$\mathbf{I}$. According to \cite{beard2012small}, the following kinematic equations apply to the positions $\mathbf{p}=[x,y,z]^T$ and the orientation $\mathbf{q}$:
\begin{equation}
    \dot{\mathbf{p}} = \mathbf{R}^{n}_{b}(\mathbf{q})\mathbf{v}
\end{equation}
\begin{equation}
    \dot{\textbf{q}} = \frac{1}{2}
    \begin{bmatrix}
        0      & -\mathbf{\omega}^T \\
        \omega & -\mathbf{S}(\mathbf{\omega}) 
    \end{bmatrix}\mathbf{q}
\end{equation}
where $\mathbf{R}^n_b$ is the rotation matrix from the body frame $\{b\}$ attached to the FWUAV center of gravity to the inertial $\{n\}$ North-East-Down (NED) world frame, while $\mathbf{v}=[u,v,w]^T$ and $\mathbf{\omega}=[p,q,r]^T$ are the linear and angular body velocities respectively and $\mathbf{S}$ is the skew-symmetric matrix.

The linear and angular velocities $\mathbf{v}$ and $\mathbf{\omega}$ expressions are derived from the Newton-Euler equations of motion:
\begin{equation}
    m\dot{\mathbf{v}} + \mathbf{\omega} \times\ \mathbf{v} = \mathbf{R}^n_b(\mathbf{q})^T m \mathbf{g}^n + \mathbf{F}_{prop} + \mathbf{F}_{aero}
\end{equation}
\begin{equation}
    \mathbf{I}\dot{\mathbf{\omega}} + \mathbf{\omega} \times \mathbf{I\omega} = \mathbf{M}_{prop} + \mathbf{M}_{aero}
\end{equation}
Where $\mathbf{g}^n$ is the force of gravity in the inertial frame. The other terms, $\mathbf{F}_{aero}$ and $\mathbf{M}_{aero}$ are the aerodynamical forces and moments while $\mathbf{F}_{prop}$ and $\mathbf{M}_{prop}$ are the propulsion forces and moments.

\subsection{Aerodynamic Forces and Moments}
\subsubsection{Wind Disturbances}
\label{subsubsec:wind-atmo-disturb}
$\mathbf{v}_r^b$ is the velocity of the FWUAV with respect to the relative wind as:
\begin{equation}
    \mathbf{v}_r^b = \mathbf{v}_g^b - \mathbf{v}_w^b
\end{equation}
with $\mathbf{v}_g^b$ the velocity vector of the aircraft relative to the ground and $\mathbf{v}_w^b$ the wind speed in the body frame. The wind speed can be decomposed into two parts: a steady ambient wind component $\mathbf{v}_{w_{s}}^n$ expressed in the inertial frame and a stochastic process $\mathbf{v}_{w_{g}}^b$ representing wind turbulence expressed in the body frame.
\begin{equation}
    \mathbf{v}_w^b = \mathbf{R}_n^b(\mathbf{q})\mathbf{v}_{w_{s}}^n + \mathbf{v}_{w_{g}}^b = 
    \begin{bmatrix}
        u_r \\
        v_r \\
        w_r \\
    \end{bmatrix}
\end{equation}
Similarly, the angular rate relative to the air comprises two terms: the nominal angular rate of the FWUAV $\mathbf{\omega}^b$ and the additional turbulent angular rates $\mathbf{\omega}_w^b$:
\begin{equation}
    \mathbf{\omega}_r^b = \mathbf{\omega}^b - \mathbf{\omega}_w^b = 
    \begin{bmatrix}
        p_r \\
        q_r \\
        r_r \\
    \end{bmatrix}
\end{equation}
The stochastic processes modeling the additional turbulence correspond to white noise passed through filters given by the Dryden spectrum model \cite{beard2012small,milspec}. The airspeed $V_a$, angle of attack $\alpha$ and sideslip angle $\beta$ are defined as:
\begin{equation}
    V_a = \sqrt{u_r^2 + v_r^2 + w_r^2}
\end{equation}
\begin{equation}
   \alpha = \tan^{-1}\left(\frac{u_r}{w_r}\right)~,~\beta = \sin^{-1}\left(\frac{v_r}{V_a}\right) 
\end{equation}
%\begin{equation}
%   \beta = \sin^{-1}\left(\frac{v_r}{V_a}\right) 
%\end{equation}

\subsubsection{Aerodynamic Model}
The flying wing Skywalker x8 is not equipped with a rudder. The available control surfaces only consist in aileron and elevator deflection angles, $\delta_a$~and~$\delta_e$, and a throttle command $\delta_t$.
The translational aerodynamic forces $\mathbf{F}_{aero}$ are the drag $D$, the lateral force $Y$ and the lift $L$. They are expressed in the wind frame and can be expressed in the body frame as:
\begin{equation}
    \mathbf{F}_{aero} = \mathbf{R}_w^b(\alpha, \beta) \begin{bmatrix} -&D \\ &Y \\ -&L \end{bmatrix}
\end{equation}
\begin{equation}
    \begin{bmatrix} D \\ Y \\ L \end{bmatrix} = \frac{1}{2} \rho V_a^2 S \begin{bmatrix} C_D(\alpha, \beta, q_r, \delta_e) \\ C_Y(\beta, p_r, r_r, \delta_a) \\ C_L(\alpha, q_r, \delta_e)\end{bmatrix}
\end{equation}
Typically, these nonlinear forces are described by $C_D, C_Y, C_L$. They consist in a set of coefficients determined by computational fluid dynamics simulations and/or experimental wind tunnel data. The expressions of the aerodynamic moments $\mathbf{M}_{aero}$ can be found following the same logic:
\begin{equation}
    \mathbf{M}_{aero} = \frac{1}{2} \rho V_a^2 S \begin{bmatrix} bC_l(\beta, p_r, r_r, \delta_a) \\ cC_m(\alpha, q_r, \delta_e) \\ bC_n(\beta, p_r, r_r, \delta_a) \end{bmatrix}
\end{equation}
where $\rho$ is the air density, $V_a$ the nominal airspeed, $S$ the wing area, $b$ the wingspan and $c$ the aerodynamic chord. The aerodynamic coefficients and geometric parameters of the x8 FWUAV are taken from~\cite{gryte2018aerodynamic}.

\subsubsection{Propulsion Forces and Moments}
The propeller of the x8 FWUAV is located at the back of the airframe and aligned with the body frame's $x$-axis such that
%\begin{equation}
   $\mathbf{F}_{prop} = [T_p,0,0]^T$.
    %\begin{bmatrix}
     %   T_p \\ 0 \\ 0\end{bmatrix}
%\end{equation}
In the current simulation, a simplified model of the engine and the propeller is used, which provides a linear relation $T_p = C \delta_t$ between the thrust force $T_p$ and the $\delta_t$ throttle command given in the $[0, 1]$ interval, with $C = 5.9$.
%\begin{equation}
%    T_p = K \delta_t
%\end{equation}

\subsection{Actuator Dynamics and Constraints}
Input commands in the simulator are normalized in the range $[-1, 1]$ for control surface deflections (${\delta_a}, {\delta_e}$) and between $[0, 1]$ for the throttle ${\delta_t}$.
Following \cite{bohn2019deep}, the aileron and elevator dynamics are described by rate-limited and saturated second-order transfer functions with natural frequency $\omega_0 = 100$ and damping $\zeta = \frac{1}{\sqrt{2}}$. The control surface deflection angles and rates are limited to $\pm 30\degree$ and $\pm 200$ $\degree$/s. The throttle dynamics are modeled by a first-order transfer function, with gain $K=1$ and time constant $T = 0.2$.

\section{CONTROL METHODS}
\label{sec:control-methods}
%In this study the controller's task is to track an attitude reference for the aforedescribed FWUAV. Attitude refers to the roll $\phi$ and pitch $\theta$ angles and the controller's role is to stabilize them to a given desired state $\phi^d$ and $\theta^d$ respectively. 
In this study the controller's task is to track desired states $\phi^d$ and $\theta^d$ for the attitude roll and pitch angles $\phi$ and pitch $\theta$ of the aforedescribed FWUAV when subject to various wind disturbances.
Following \cite{bohn2023data}, the airspeed remains controlled by a PI controller driving the throttle command $\delta_t$ to maintain a nominal airspeed, with the gains tuned as in~\cite{bohn2019deep}: $k_{p_{V}} = 0.5$ and $k_{i_{V}} = 0.1$.
This airspeed PI controller is common to all the following controllers. This choice was motivated by the fact that when certain references are generated, the desired pitch can contradict the desired airspeed, e.g. low desired airspeed and nose down desired pitch. In a case where one would control the desired altitude, the outer loop would give coherent desired states to the inner attitude control loops, but this is beyond the scope of the present work.

\subsection{PID Control}
PID control is a widely used controller for UAV attitude control because of its ease of use and is therefore considered as a reference point in the RL literature for UAV control \cite{bohn2019deep,bohn2023data,koch2019reinforcement,lambert2019low}. We fine-tuned the PID gains given by \cite{bohn2019deep} to better suit our simulated loops in nominal conditions i.e. with no wind disturbances, with the gains presented in Table \ref{tab:pidrollpitch}.
\begin{table}[h]
\centering
\caption{PID controller parameters}
\begin{tabular}{|cc|cc|}
\hline
Parameter & Value & Parameter & Value \\
\hline
    $k_{p\phi}$ & $1.5$ & $k_{p\theta}$ & $-2$ \\
    $k_{i\phi}$ & $0.1$ & $k_{i\theta}$ & $-0.3$ \\
    $k_{d\phi}$ & $0.1$ & $k_{d\theta}$ & $-0.1$ \\
\hline
\end{tabular}
\label{tab:pidrollpitch}
\end{table}

\subsection{Reinforcement Learning (RL) Control}
\label{subsubsec:method-base-rl}
In RL, the control problem is formulated as that of finding an optimal action strategy in an environment modeled as a Markov Decision Process (MDP) \cite{sutton2018reinforcement} comprised of:
\begin{itemize}
    \item A set of actions $a \in A$, here the same as in the roll and pitch PID loops with the superscript $c$ denoting a commanded position of the aileron and elevator, $a = ({\delta^c_a}$, ${\delta^c_e})$.
    \item A set of states $s \in S$, based on \cite{bohn2023data}:
    \begin{itemize}
        \item the attitude angles: roll $\phi$ and pitch $\theta$;
        \item the airspeed $V_a$;
        \item the angular rates $\omega_r^b = [p_r, q_r, w_r]^T$;
        \item the angle of attack $\alpha$ and the angle sideslip $\beta$;
        \item the roll and pitch errors:
            \begin{equation}
                \begin{aligned}
                    e_\phi &= \phi^d - \phi \\
                    e_\theta &= \theta^d - \theta
                \end{aligned}
            \end{equation}
        \item the roll and pitch integral errors:
            \begin{equation}
                I_{e_{*}} = I_{e_{*}} + e_* \cdot dt, \: * = \{\phi, \theta\}, dt=0.01
            \end{equation}
            where $dt$ is the simulation period, also equal to the control rate. The integral error is reset at the beginning of each episode or target change.
        \item The last action taken by the agent: ${\delta_{a|_{t-1}}^c}$ and ${\delta_{e|_{t-1}}^c}$.
    \end{itemize}
    \item A transition function $T$, defining a distribution over the next states given a current state and action: $P(s_{t+1} | s_t, a_t) = T(s_t, a_t)$. 
    \item A scalar reward function $R_t$, outputting a score, rewarding the agent for good actions and penalizing bad ones, that shapes the agent's task.
\end{itemize}

\noindent The reward function has been adapted from \cite{bohn2019deep} as follows:
\begin{equation}
    \begin{aligned}
        R_{\phi} &= \text{clip}\left(\frac{| \phi - \phi^d |}{\zeta_\phi}, 0, c_{\phi}\right) \\
        R_{\theta} &= \text{clip}\left(\frac{| \theta - \theta^d |}{\zeta_\theta}, 0, c_\theta \right) \\
        R_t &= -(R_\phi + R_\theta) \\
        \zeta_\phi &= 3.3,\:\zeta_\theta = 2.25,\:c_\phi=c_\theta=0.5
    \end{aligned}
    \label{eq:base-reward}
\end{equation}
where $\zeta_*$ are scaling coefficients to take into account the differences between roll and pitch error ranges, each reward component is clipped to 0.5 so the minimal total reward equals to -1.

RL algorithms aim to discover the best mapping $\pi$ from a state to an action (control law), in the RL literature, referred as the optimal policy ($\pi^*$), and defined as the policy leading to trajectories $\tau = (s_0, a_0, s_1, a_1,...)$ that maximize the expected discounted sum of accumulated rewards:
\begin{equation}
\pi^* = \argmax_{\pi} \mathbb{E}_{\tau \sim \pi} \left[\sum_{t=0}^{\infty} \gamma^t R_t(s_t, a_t)\right]
\end{equation}
with $\gamma$ being the discount factor.

This policy, often represented by an artificial neural network (ANN) with learnable parameters $\Theta$, serves as a nonlinear controller.

\subsubsection{Model-Free RL: PPO \& SAC}
We choose Proximal Policy Optimization (PPO) \cite{schulmanPPO} and Soft Actor-Critic (SAC) \cite{haarnoja2018sac} to represent the MF-RL end of the control spectrum because of their state-of-the-art performance for robotic applications with continuous action spaces. Moreover, SAC's choice was motivated by its similar policy update to the mixed method used in this work (TD-MPC).

% \begin{figure}[h!]
%     \centering
%     \includegraphics[width=0.45\textwidth]{img/mfrl_diagram.pdf}
%     \caption{\textbf{MF-RL Control Block Diagram.} The agent is a neural network directly outputting an action (policy).}
%     \label{fig:mfrl-ctl-blk}
% \end{figure}

\subsubsection{Model-Based Method: TD-MPC}
Temporal Difference Model Predictive Control (TD-MPC) \cite{Hansen2022tdmpc} is an MB-RL, mixed control method combining an explicit dynamics model and a terminal value function learned by using Temporal Difference learning in an RL setting. We use its latest implementation from \cite{hansen2024tdmpc2}.

\textbf{Trainable Models}. TD-MPC learns a Task-Oriented Latent Dynamics (TOLD) model, comprised of several MLPs (Multi-Layer Perceptrons) with trainable parameters $\Theta$. They aim at modeling certain functions, most notably: the agent's dynamics, the reward function $R_\Theta$, the Q-value function $Q_\Theta$ and the policy. During training, a trajectory of prediction horizon $H$ is sampled from a replay buffer in an off-policy RL fashion. Using this sampled trajectory, the TOLD model is trained by optimizing a loss function including: the reward prediction function, the dynamics model and a Bellman error component for $Q_\theta$. The policy is learned by maximizing a similar objective to the SAC algorithm \cite{haarnoja2018sac}.

\textbf{Planning}. TD-MPC uses a control-theory basis for planning, utilizing Model Predictive Path Integral (MPPI) \cite{williams2015MPPI}. MPPI is a sampling-based model-predictive control (MPC) algorithm iteratively learning the parameters of a multivariate Gaussian distribution using importance sampling. 
First, TD-MPC simulates candidate trajectories $\Gamma$ by sampling a mixture of trajectories of horizon $H$ by querying both the Gaussian distribution and the learned policy function $\pi_\Theta$ for actions and by recursively predicting next states using the learned dynamics model. One of TD-MPC's key idea is to use the learned Q-value function $Q_\Theta$ to give an estimation of the terminal value when computing a trajectory's total return $\phi_\Gamma$. Here, $Q_\Theta$ gives an estimate of the value of the trajectory beyond the horizon limit, emulating infinite-horizon MPC.
\begin{equation}
    \phi_\Gamma \triangleq \mathbb{E}_{\Gamma} [  \underbrace{\gamma^H Q_{\Theta}(z_H, a_H)}_{\substack{\text{Terminal value} \\ \text{estimated by\: $Q_\Theta$}}} + \underbrace{\sum_{t=0}^{H-1} \gamma^t R_{\Theta}(z_t, a_t)}_{\substack{\text{Total return of the} \\ \text{candidate trajectory over} \\ \text{the time horizon } H}} ]
\end{equation}

\begin{figure}[h!]
    \centering
    \includegraphics[width=0.45\textwidth]{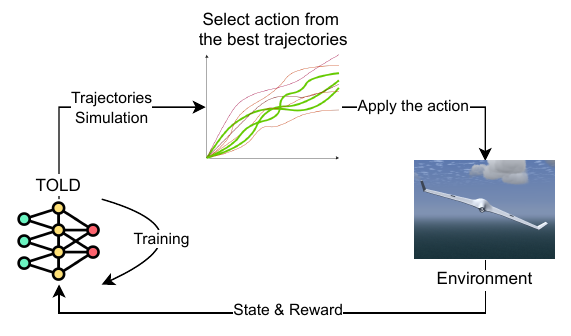}
    \caption{\textbf{TD-MPC Control Block Diagram.} 1) The trained TOLD model is used for simulating candidate trajectories and estimating their return. 2) Using importance sampling update an action Gaussian distribution and sample it. 3) Sample an action from the action distribution and apply it.}
    \label{fig:tdmpc-ctl-blk}
\end{figure}

As illustrated in Figure \ref{fig:tdmpc-ctl-blk}, after taking the trajectories with the best return, MPPI computes the action Gaussian parameters update by using importance sampling, see~\cite{Hansen2022tdmpc}. The action Gaussian is then sampled and the first action is sent to the actuators, as in receding-horizon MPC.

% \begin{equation}
%     \begin{aligned}
%         \phi_\Gamma \triangleq \mathbb{E}_{\Gamma} \left[  \underbrace{\gamma^H Q_{\theta}(z_H, a_H)}_{\text{Terminal value estimated by}\:Q_\theta} \right.\\ 
%         + \left. \underbrace{\sum_{t=0}^{H-1} \gamma^t R_{\theta}(z_t, a_t)}_{\substack{\text{Total return of the simulated} \\ \text{trajectory over the time horizon H}}} \right]
%     \end{aligned}
% \end{equation}

\section{EXPERIMENTAL SETUP}
\label{sec:exp-setup}
\subsection{Simulation}
We conducted our experiments in the JSBSim flight simulator \cite{berndt2004jsbsim} because of its ability to handle fixed-wing aircraft flight dynamics and to model all the studied wind disturbances: constant wind, wind gusts and stochastic turbulences. JSBSim was preferred over the PyFly simulator presented in~\cite{bohn2019deep} because of its wide use in the aerospace open-source and research communities and its built-in wind gust generation \cite{moallemi2016b,mathisen2021precision,de2023deep}.

We are interested in separating the performance of the studied control methods between linear and nonlinear regimes. Therefore, we introduce a classification of attitude angle references in Table \ref{tab:targetranges}, with nominal and hard levels representing linear and nonlinear regions of the state space respectively.
\begin{table}[h]
\small{
\fontsize{8}{9}
\centering
\caption{References classification}
    \begin{tabular}{|c|cc|}
    \hline
    Difficulty & Roll(°) & Pitch(°) \\
    \hline
        Nominal & $[-45, 45]$  &  $[-25, 25]$ \\
        Hard    & $[-60, -45]\cup[45, 60]$ & $[-30, -25]\cup[25, 30]$ \\
    \hline
    \end{tabular}
    \label{tab:targetranges}
}
\end{table}

\subsection{Implementation Details}
For the optimization algorithms, we used the CleanRL PPO \& SAC implementations \cite{huang2022cleanrl} with the parameters presented in Tables \ref{tab:ppoparams} \& \ref{tab:sacparams}. We used the most recent implementation of TD-MPC, TD-MPC2 \cite{hansen2024tdmpc2} with the parameters specified in Table \ref{tab:tdmpcparams}.

We trained the RL agents for 375 episodes of 2000 timesteps each, i.e. 20 seconds of flight time at 100~Hz. At the start of every episode, we initialize the FWUAV  at 600~m above sea level, with a nominal speed of 17 m/s and roll, pitch, yaw $\{\phi, \theta, \psi\} = 0$. We compute the mean and standard error of the mean over 5 runs for each metric. Training was conducted on a RTX A6000 with 48GB of memory.

\begin{table}[h]
\small{
\centering
\vspace{0.3cm}
\caption{PPO Parameters.}
\begin{tabular}{|cc|cc|}
\hline
Parameter & Value & Parameter & Value \\
\hline
    LR               & 3e-4    & Entropy coef          & 1e-2 \\
    Parallel envs               & $6$       & Value fn coef   & $0.5$ \\
    Rollout steps               & $2048$    & Total timesteps  & 750k \\
    Discount factor $\gamma$    & $0.99$    & Minibatches    & $32$ \\
    GAE $\lambda$               & $0.95$     & Batch size   &  $12 228$ \\
    PPO clip                    & $0.2$ & & \\
\hline
\end{tabular}
\label{tab:ppoparams}
}
\end{table}

\begin{table}[h]
\small{
\fontsize{7}{7}
\centering
\vspace{0.3cm}
\caption{SAC Parameters.}
\begin{tabular}{|cc|cc|}
\hline
Parameter & Value & Parameter & Value \\
\hline
    Buffer size               & 1e6    & Q-net LR & 1e-3 \\
    Discount factor $\gamma$  & 0.99       & $\pi$ update freq  & 2 \\
    $Q_{targ}$ smoothing $\tau$ & 5e-3    &  $Q_{targ}$ update freq & 1 \\
    Batch size & 256    & Total timesteps   & 750k \\
    Seed steps & 5000     & Noise clip & 0.5 \\
    Policy LR  & 3e-4 & Entropy reg & 0.2 \\
\hline
\end{tabular}
\label{tab:sacparams}
}
\end{table}

\begin{table}[h]
\small{
\centering
\vspace{0.3cm}
\caption{TD-MPC Parameters.}
\begin{tabular}{|cc|cc|}
\hline
Parameter & Value & Parameter & Value \\
\hline
    Buffer size               & 1e6     & Total timesteps   & 750k \\
    Discount factor $\gamma$  & 0.99    & Temporal coef     & 0.5  \\
    Target smoothing $\tau$ & 0.01      & MPPI Iterations    & 6  \\
    Batch size              & 256       & MPPI Samples      & 512 \\
    Seed steps              & 10 000     & MPPI Elites      & 64 \\
    LR                      & 3e-4       & MPPI $\pi$ Trajs    & 24\\
    Reward coef             & 0.1        & Horizon          & 3 \\
    Value coef              & 0.1       & Temperature       & 0.5\\
    Consistency coef        & 20        &  & \\
\hline
\end{tabular}
\label{tab:tdmpcparams}
}
\end{table}

\section{RESULTS AND DISCUSSIONS}
\label{sec:results}

Our first experiment aims at comparing RL agents in nominal, non-disturbed conditions and with respect to reference tracking only (Section \ref{subsubsec:res-base-rl}).
We also show that RL agents tend to converge towards policies that produce oscillating actions, and we evaluate two strategies to mitigate the issue (Section \ref{subsubsec:res-act-oscill}).
Then we study the impact of turbulence and wind gusts separately on each algorithm performances (Section \ref{subsec:res-atmo-disturb}) and provide a deeper discussion of the results (Section\ref{res:discussions}).

%First, as reported in (Section \ref{subsubsec:res-base-rl}), we focus on a simple nominal environment without any wind disturbance. The interest is in studying the behavior of RL agents and their potential benefits. In Section \ref{subsubsec:res-act-oscill} we observe that RL agents tend to converge towards policies that produce oscillating actions and introduce actuation fluctuation as a key metric for our comparison, we thus present methods to mitigate such policies and provide results on their efficiency to lower their actuation fluctuation. 

%Second, as described in Sections \ref{subsubsec:res-stochastic-turb} and \ref{subsubsec:res-wind-gusts}, we focus on adding stochastic turbulence and wind gusts, respectively, as additional sources of complexity.

\subsection{Nominal Wind Conditions}
\label{subsec:res-nominal-atmo}
This section focuses on evaluating all 4 controllers under nominal wind conditions i.e. no wind, no turbulence and no wind gusts. Agents are trained and tested with these wind settings and across the two reference difficulty levels from Table~\ref{tab:targetranges}.

\subsubsection{Base RL Algorithms}
\label{subsubsec:res-base-rl}

% \begin{table*}[h!]
%     \centering
%     \begin{tabular}{|c|c|c|c|}
%         \hline
%         \rowcolor{gray!25}Agent & Nominal Refs & Hard Refs \\
%         \cline{2-3}
%         \hline
%         \cellcolor{gray!25} PID & 0,0420 & 0,2010\\
%         \hline
%         \cellcolor{blue!25}TD-MPC & \cellcolor{blue!25}0,0381 $\pm$ 0.0010 & \cellcolor{blue!25}0,0910 $\pm$ 0,0069 \\
%         \hline
%         \cellcolor{gray!25} PPO & 0,0529 $\pm$ 0,0012 & 0,1578 $\pm$ 0,0248 \\
%         \hline
%         \cellcolor{gray!25} SAC & 0,0664 $\pm$ 0,0083  & 0,1115 $\pm$ 0,0065 \\
%         \hline
%     \end{tabular}
%     \caption{\textbf{Nominal Atmosphere.} Average of roll and pitch tracking RMSE, over all reference difficulty levels. Best scores are highlighted in blue.}
%     \label{tab:res-nominal-noactreg}
% \end{table*}

\begin{table*}[h!]
    \centering
    \begin{tabular}{|c|c|c|}
        \hline
        \rowcolor{gray!25}Agent & Nominal Refs {\footnotesize(x$10^{-2}$)} & Hard Refs {\footnotesize(x$10^{-2}$)}\\
        \hline
        \cellcolor{gray!25} PID & 4.20 & 20.10\\
        \hline
        \cellcolor{blue!25}TD-MPC & \cellcolor{blue!25}3.81 {\footnotesize $\pm$ 0.10} & \cellcolor{blue!25}9.10 {\footnotesize $\pm$ 0.69} \\
        \hline
        \cellcolor{gray!25} PPO & 5.29 {\footnotesize $\pm$ 0.12} & 15.78 {\footnotesize $\pm$ 2.48} \\
        \hline
        \cellcolor{gray!25} SAC & 6.64 {\footnotesize $\pm$ 0.83}  & 11.15 {\footnotesize $\pm$ 0.65} \\
        \hline
    \end{tabular}
    \caption{\textbf{Nominal Atmosphere.} Average of roll and pitch tracking RMSE, over all reference difficulty levels. Best scores are highlighted in blue.}
    \label{tab:res-nominal-noactreg}
\end{table*}

\begin{figure*}[th!]
    \centering
    \begin{subfigure}[t]{\textwidth}
        \centering
        \includegraphics[width=0.95\textwidth]{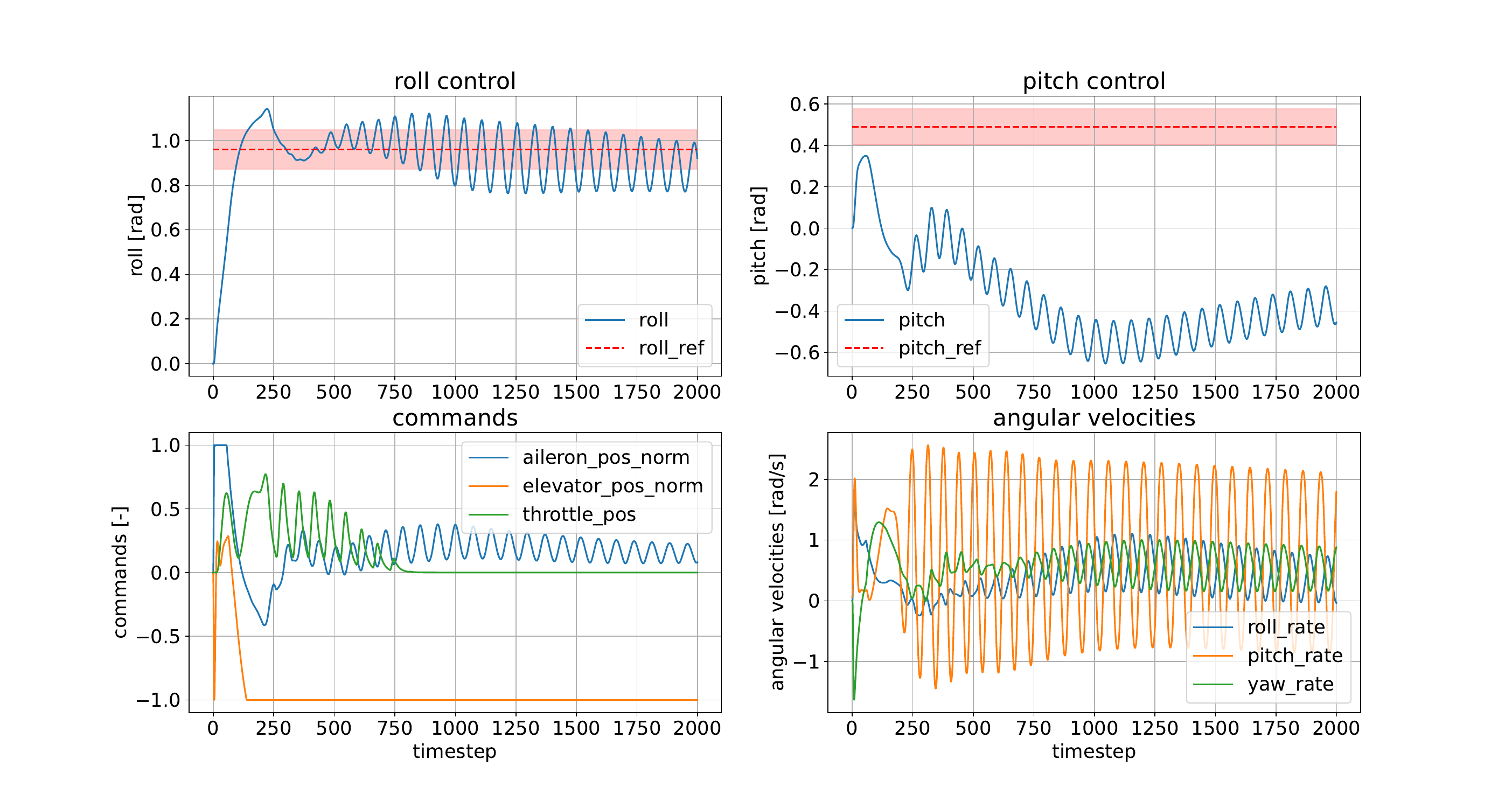}
        \caption{PPO}
    \end{subfigure}
    \begin{subfigure}[t]{\textwidth}
        \centering
        \includegraphics[width=0.95\textwidth]{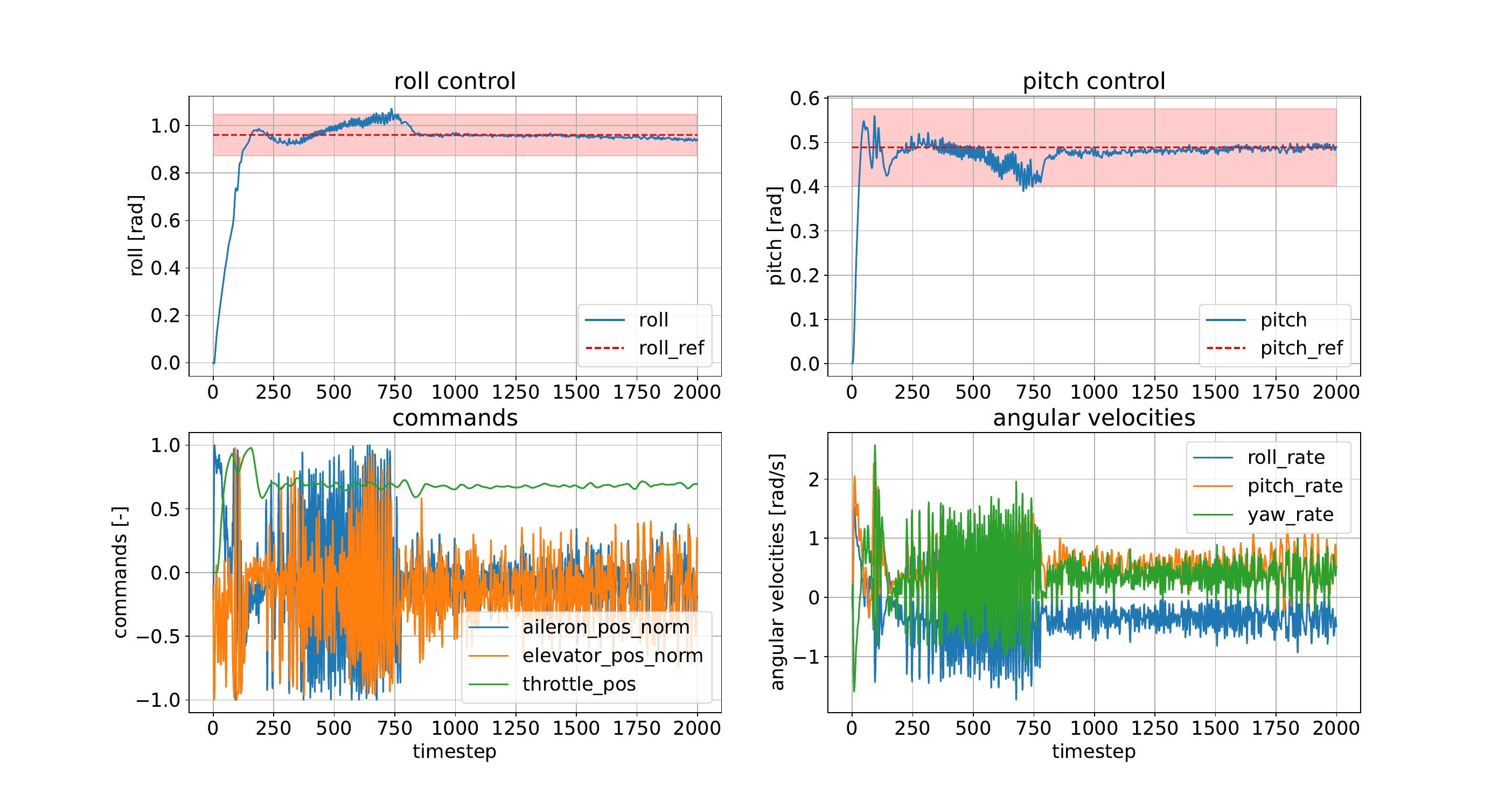}
        \caption{TD-MPC}
    \end{subfigure}
    \caption{\textbf{TD-MPC's Superiority for Hard References: PPO vs TD-MPC}. Red dashed lines are the references: Roll = $55\degree$, Pitch = $28\degree$. The red area around the reference line corresponds to $\pm 5\degree$ error bounds}
    \label{fig:no-atmo-ppovstdmpc}
\end{figure*}

%As expected, 
The results from Table \ref{tab:res-nominal-noactreg} confirm that PID is a strong baseline in nominal conditions, with only TD-MPC from RL algorithms performing slightly better.
On the contrary, under more nonlinear dynamics forced by hard references, all RL methods clearly outperform PID, with a significant advantage of TD-MPC over PPO and SAC. 
Figure~\ref{fig:no-atmo-ppovstdmpc} illustrates TD-MPC's better hard reference tracking compared to PPO.
Overall TD-MPC displays moderate to strong gains over all methods across both dynamics regimes.

%Table \ref{tab:res-nominal-noactreg} highlights the good level of performance of PID in the linear part of the state space (nominal references) but also its limitations under nonlinear dynamics (hard references).
%Model-free agents (PPO and SAC) present good RMSE score on nominal references while also doing better on hard references than PID, displaying better ability to capture nonlinear dynamics than PID. Table \ref{tab:res-nominal-noactreg} consistently shows better RMSE scores for TD-MPC across all reference difficulties. While giving a modest advantage on nominal references, the performance gap is striking for hard references, where the dynamics of the FWUAV becomes nonlinear. 

\subsubsection{RL Policies and High Action Oscillation}
\label{subsubsec:res-act-oscill}

\begin{table*}[h!]
    \centering
        \begin{tabular}{|c|c|c|c|c|}
        \hline
        \rowcolor{gray!25}\multirow{2}{*}{} & \multicolumn{2}{|c|}{Nominal Refs {\footnotesize(x$10^{-2}$)}} & \multicolumn{2}{|c|}{Hard Refs {\footnotesize(x$10^{-2}$)}} \\
        \cline{2-5}
        \hline
        \cellcolor{gray!25}Agent & \cellcolor{gray!25}RMSE & \cellcolor{gray!25}Act Fluct & \cellcolor{gray!25}RMSE & \cellcolor{gray!25}Act Fluct \\
        \hline
        \cellcolor{red!25}PID & 4.20 & \cellcolor{red!25}0.07 & 20.10 & \cellcolor{red!25}0.17 \\
        \hline
        \cellcolor{gray!25}TD-MPC & 3.81 {\footnotesize $\pm$ 0.10} & 4.37 {\footnotesize $\pm$ 0.48} & 9.10 {\footnotesize $\pm$ 0.69} & 6.02 {\footnotesize $\pm$ 0.59}\\
        \cellcolor{blue!25}TD-MPC AVP & \cellcolor{blue!25}\textbf{3.58 {\footnotesize $\pm$ 0.01}} & \textbf{0.54 {\footnotesize $\pm$ 0.02}} & \cellcolor{blue!25}\textbf{8.09 {\footnotesize $\pm$ 0.03}} & \textbf{0.60 {\footnotesize $\pm$ 0.02}} \\
        \hline
        \cellcolor{gray!25} PPO & 5.29 {\footnotesize $\pm$ 1.20} & 1.98 {\footnotesize $\pm$ 0.58} & 15.78 {\footnotesize $\pm$ 2.48} & 5.54 {\footnotesize $\pm$ 0.93}\\
        \cellcolor{gray!25}PPO CAPS & \textbf{5.05 {\footnotesize $\pm$ 1.50}} & \textbf{1.47 {\footnotesize $\pm$ 0.27}} & 19.17 {\footnotesize $\pm$ 4.68} & \textbf{4.56 {\footnotesize $\pm$ 0.60}} \\
        \cellcolor{gray!25}PPO AVP & 5.78 {\footnotesize $\pm$ 2.10} & 13.78 {\footnotesize $\pm$ 3.16} & \textbf{13.80 {\footnotesize $\pm$ 0.82}} & 13.48 {\footnotesize $\pm$ 2.50} \\
        \hline
        \cellcolor{gray!25} SAC & 6.64 {\footnotesize $\pm$ 0.83} & 10.07 {\footnotesize $\pm$ 1.12} & \textbf{11.15 {\footnotesize $\pm$ 0.65}} & 12.00 {\footnotesize $\pm$ 0.71}\\
        \cellcolor{gray!25}SAC CAPS & 6.79 {\footnotesize $\pm$ 0.74} & 0.53 {\footnotesize $\pm$ 0.27} & 12.28 {\footnotesize $\pm$ 1.17} & 1.44 {\footnotesize $\pm$ 0.55} \\
        \cellcolor{gray!25}SAC AVP & \textbf{6.54 {\footnotesize $\pm$ 0.75}} & \textbf{0.50 {\footnotesize $\pm$ 0.18}} & 15.14 {\footnotesize $\pm$ 3.39} & \textbf{1.20 {\footnotesize $\pm$ 0.32}} \\
        \hline
    \end{tabular}
    \caption{\textbf{Nominal Atmosphere + Actuation Regulation (through AVP or CAPS).} Average of roll - pitch tracking errors and aileron - elevator action fluctuation, over all reference difficulty levels. \textbf{Bold} values are the best score among one agent type. \colorbox{blue!25}{Blue}: best RMSE score across all agents. \colorbox{red!25}{Red}: best Actuator Fluctuation score across all agents.}
    \label{tab:res-nominal-actreg}
\end{table*}

The RL literature usually focuses on performance metrics only \cite{koch2019reinforcement,Hansen2022tdmpc}, but for true robotic applications, we practitioners must also make sure to minimize wear and tear as well as energy consumption. 
This can only be done by monitoring actuation solicitation and thus we incorporate in our evaluation an actuation fluctuation (Act Fluct) metric, following \cite{song2023lipsnet}:
%An important feature robotics controllers should embed is energy consumption, which is directly related to actuators solicitation. Common RL literature in simulated environments eludes actuation fluctuation and focuses on pure performance of agents similar to the results reported in Section~\ref{subsubsec:res-base-rl}. Hence, a new actuation fluctuation (Act Fluct) metric is introduced in this section, following \cite{song2023lipsnet}:
\begin{equation}
    \xi(\pi) = \frac{1}{2} \sum_{j \in [a,e]} \frac{1}{E} \sum_{n=1}^{E} \left[ \frac{1}{T} \sum_{t=1}^{T} |\delta_{j|_{t}} - \delta_{j|_{t-1}}| \right]
\end{equation} 
It incentivizes the minimization of the distance between two consecutive actuator positions ($\delta_a$ or $\delta_e$).

Table \ref{tab:res-nominal-actreg} shows that high performances of RL methods come at the cost of strongly oscillating actions.
%Despite converging to flying policies, Table \ref{tab:res-nominal-actreg} displays highly oscillating actions for most RL policies.
In fact, RL can lead to bang-bang control, where the learned optimal policy consists in alternating abruptly between actions far from each other in the action space \cite{seyde2021bang}.
To mitigate this issue, we evaluate two oscillation regulation methods from the literature.
%This ablation studies the effects of two action regulation methods: Action Variation Penalty (AVP) in the reward and Conditioning for Action Policy Smoothness (CAPS) \cite{mysore2021regularizing}.

\textbf{Action Variation Penalty (AVP)}. Adding an AVP in the reward function is a common way of addressing the bang-bang policy problem. Hence, we modify the base reward of Equation \ref{eq:base-reward} and propose adding a simple additional reward component $R_\delta$, similar to \cite{bohn2019deep} for penalizing abrupt changes of consecutive actions, forming the reward function denoted $R_{AVP}$:
\begin{equation}
    \begin{aligned}
        R_{\delta} &= \text{clip}\left(\frac{\frac{1}{2} \sum_{j \in [a, e]} |{\delta^c}_{j|_t} - {\delta^c}_{j|_{t-1}} |}{\zeta_\delta}, 0, c_\delta \right) \\ {\delta_{a|_{t-1}}^c}
        R_{AVP} &= -(R_\phi + R_\theta + R_\delta) \\
        \zeta_\phi &= 3.3, \zeta_\theta = 2.25, \zeta_\delta = 2
    \end{aligned}
    \label{eq:reward1}
\end{equation}
We separately tuned the $c_*$ clipping coefficients of the AVP reward function for each RL method, giving $c_\phi = c_\theta = 0.25$ and $c_\delta = 0.5$ for TD-MPC and SAC. However, such a tuning led PPO not converging to a successful flying policy, therefore we found an appropriate tuning for this method as $c_\phi = c_\theta = 0.45$ and $c_\delta = 0.1$. %, outputting a flying policy. 

\textbf{Conditioning for Action Policy Smoothness (CAPS).} Another candidate to mitigate this behavior is the Conditioning for Action Policy Smoothness (CAPS) loss \cite{mysore2021regularizing}. CAPS is added to the policy loss and prevents the emergence of non-smooth action policies by applying temporal and spatial regularization:
\begin{equation}
    \begin{aligned}
        L_{\text{CAPS}}(\pi_\Theta) &= \lambda_{\text{TS}} \cdot L_{\text{TS}}(\pi_\Theta) + \lambda_{\text{SS}} \cdot L_{\text{SS}}(\pi_\Theta) \\
        L_{\text{TS}}(\pi_\Theta) &= \| \pi_{\Theta}(s_t) - \pi_{\Theta}(s_{t+1}) \|_2 \\
        L_{\text{SS}}(\pi_\Theta) &= \| \pi_{\Theta}(s_t) - \pi_{\Theta}(\hat{s}_t) \|_2, \quad \hat{s}_t \sim \mathcal{N}(s_t, 0.01)
    \end{aligned}
\end{equation}
The time-only version of the CAPS loss is used to enforce temporal smoothness (with $\lambda_{TS} = 0.05$) while leaving room for high reactivity in the state space. 

\textbf{Results.} Table \ref{tab:res-nominal-actreg} exhibits the consistent improvement of the action fluctuation metric across all MF-RL agents with CAPS.
For RMSE tracking performance, Table \ref{tab:res-nominal-actreg} confirms that TD-MPC remains better than PID and MF-RL agents.

\begin{figure*}[h!]
    \centering
    \begin{subfigure}[t]{\textwidth}
        \centering
        \includegraphics[width=0.95\textwidth]{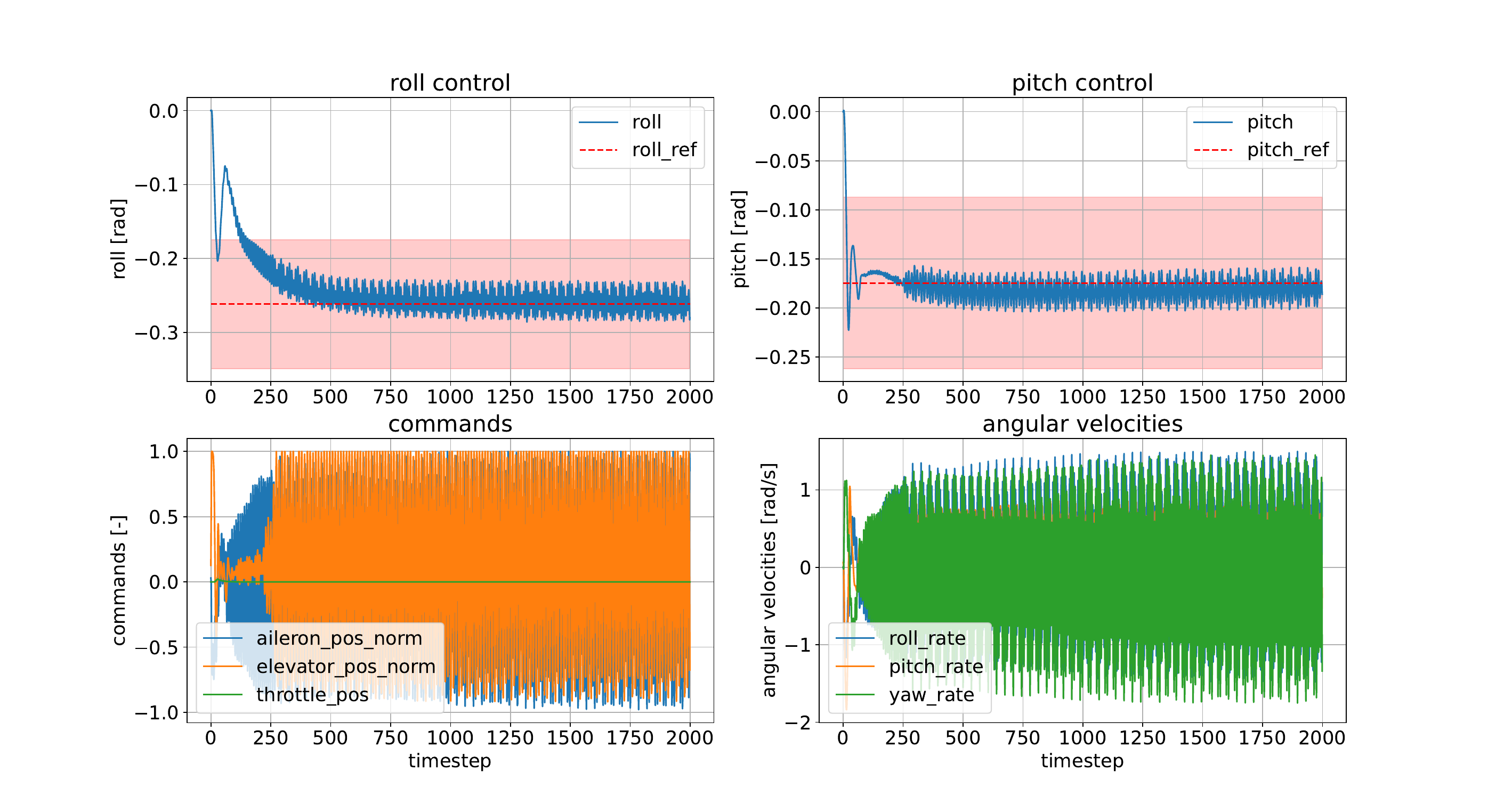}
        \caption{Base SAC}
    \end{subfigure}
    \begin{subfigure}[t]{\textwidth}
        \centering
        \includegraphics[width=0.85\textwidth]{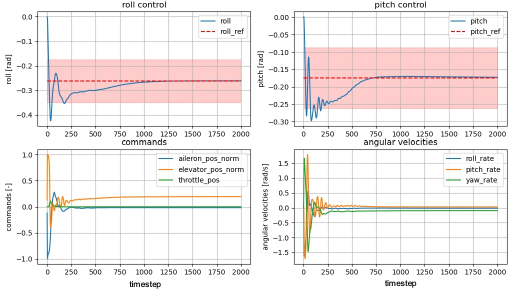}
        \caption{SAC + CAPS}
    \end{subfigure}
    \caption{\textbf{Mitigating Highly Action Oscillating Policy: SAC vs SAC + CAPS}. Red dashed lines are the references: Roll = $-15\degree$, Pitch = $-10\degree$. The red area around the reference line corresponds to $\pm 5\degree$ error bounds}
    \label{fig:no-atmo-actreg-perfs}
\end{figure*}

As for action regulation methods, we observe that action regulation methods remain algorithm dependent. Table \ref{tab:res-nominal-actreg} shows that the AVP reward function has a strong effect on actuation fluctuation, reducing it by at least an order of magnitude for TD-MPC and SAC. Conversely, AVP for PPO shows a drastic increase in actuation fluctuation, which highlights the difficult task of tuning the reward function with the goal of combining a lower actuator fluctuating flying policy. In fact, one tuning of AVP can work for certain algorithms such as SAC and TD-MPC and give an underperforming PPO agent requiring a specific tuning. CAPS appears to be a better action regulation method for MF-RL methods because of its consistency with a single tuned parameter $\lambda_{TS} = 0.05$. As an example, Figure \ref{fig:no-atmo-actreg-perfs} outlines CAPSs' action smoothing for SAC through a trajectory plot. As a result, the CAPS method has been retained for the following ablations. Despite showing great results for MF-RL methods, CAPS is not directly applicable to TD-MPC because it only regularizes its learned policy prior, while TD-MPC never directly outputs an action by sampling this prior. Indeed, we found in additional experiments that applying CAPS led to learning a smooth action policy prior but did not guarantee smooth actions after the planning step.

\subsection{Wind Disturbances}
\label{subsec:res-atmo-disturb}
Aerial robotics present challenging environmental conditions which add a layer of complexity for control systems. This section focuses on studying the impact of such conditions on our four controllers of interest. Wind disturbances are divided into two subcategories: stochastic turbulence modeled by probabilistic dynamics and wind gusts consisting in unpredictable and sudden changes in dynamics. Agents are tested and trained under stochastic turbulence in Section~\ref{subsubsec:res-stochastic-turb} and wind gusts in Section \ref{subsubsec:res-wind-gusts} separately. 

The parameters for turbulence and gusts are drawn uniformly along the values reported in Table \ref{tab:atmosettings}. This classification of constant winds is only used for evaluation. During training, constant wind magnitudes are uniformly sampled from the continuous interval $[0, 23]$ m/s. Constant wind and wind gusts directions with respect to the NED inertial frame are also sampled uniformly in the $[-1, 1]$ interval. 

To better isolate the effects of each disturbance from the effects of hard attitude angle reference tracking, the agents were trained and tested only for the nominal attitude reference angle range from Table~\ref{tab:targetranges}.

\begin{table}[h]
    \centering
    \caption{\textbf{Wind Disturbance Severity Levels (m/s).} The classification for constant wind and gust magnitudes is taken from \cite{bohn2019deep}. "Turbulence W20" is a predefined parameter in JSBSim Dryden model turbulence classification~\cite{berndt2004jsbsim}}.
    \small{
    \begin{tabular}{|c|ccc|} \hline 
         &  Constant Wind&  Turbulence W20&  Gust\\ \hline 
         OFF & 0 & 0 & 0 \\ 
         Light&  7&  7.6&  7\\  
         Moderate& 15& 15.25& 15\\  
         Severe& 23& 22.86& 23\\ 
         \hline
    \end{tabular}
    }
    \label{tab:atmosettings}
\end{table}

\subsubsection{Stochastic Turbulence}
\label{subsubsec:res-stochastic-turb}
\begin{figure}[h!]
    \centering
    \includegraphics[width=0.45\textwidth]{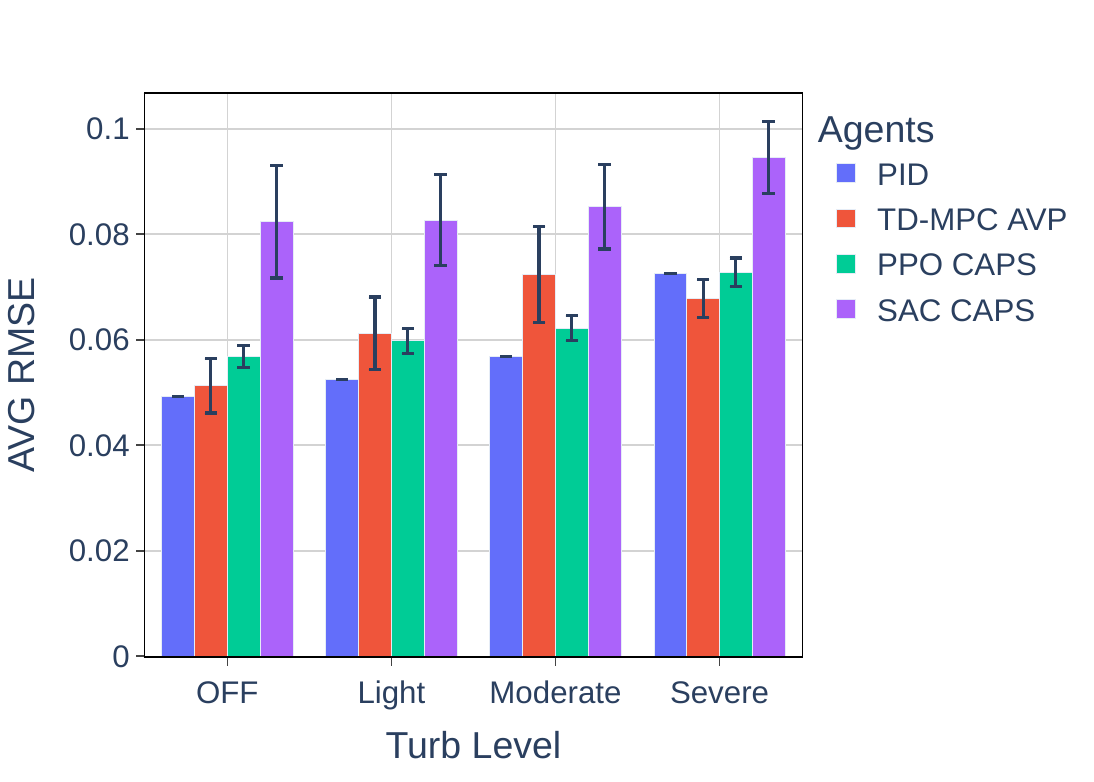}
    \caption{\textbf{Stochastic Turbulences + Action Regulation} Average tracking RMSE on nominal reference difficulty.}
    \label{fig:res-turb-barplot}
\end{figure}

As explained in Section \ref{subsubsec:wind-atmo-disturb}, turbulence is simulated as a stochastic process applying linear and angular forces to the FWUAV.
On one side, Figure \ref{fig:res-turb-barplot} outlines SAC's underperforming across all turbulence levels. On the other side, TD-MPC shows good performance, reaching the same levels of performance as PID and PPO. We observe that the severity of wind disturbance dominates over the choice of algorithm to predict the final performances.
In contrast to Section \ref{subsubsec:res-base-rl}, where TD-MPC showed a clear advantage on deterministic nonlinear dynamics for tracking high attitude reference angles, here its learned dynamics model seems here to have difficulty capturing the stochastic turbulent dynamics of this scenario, and can only resort to learning an ``on-average" robust policy. To the best of our knowledge, TD-MPC has not been applied to environments with stochastic dynamics and adapting TD-MPC to stochastic environments remains an open problem.

\subsubsection{Wind Gusts}
\label{subsubsec:res-wind-gusts}
In our simulated environment, wind gusts are triggered twice per episode at timesteps 500 and 1500. Gusts are parameterized with a startup duration (time for reaching maximum wind speed) of 0.25~s, a steady duration (time span where the wind stays at maximal magnitude) of 0.5~s, and an end duration (time for the gust to disappear) of 0.25~s. The startup and end transients are modeled as a smooth cosine function. The magnitude level of the gust is determined by its maximum wind speed and follows Table \ref{tab:atmosettings}.

\begin{figure}[h!]
    \centering
    \includegraphics[width=0.45\textwidth]{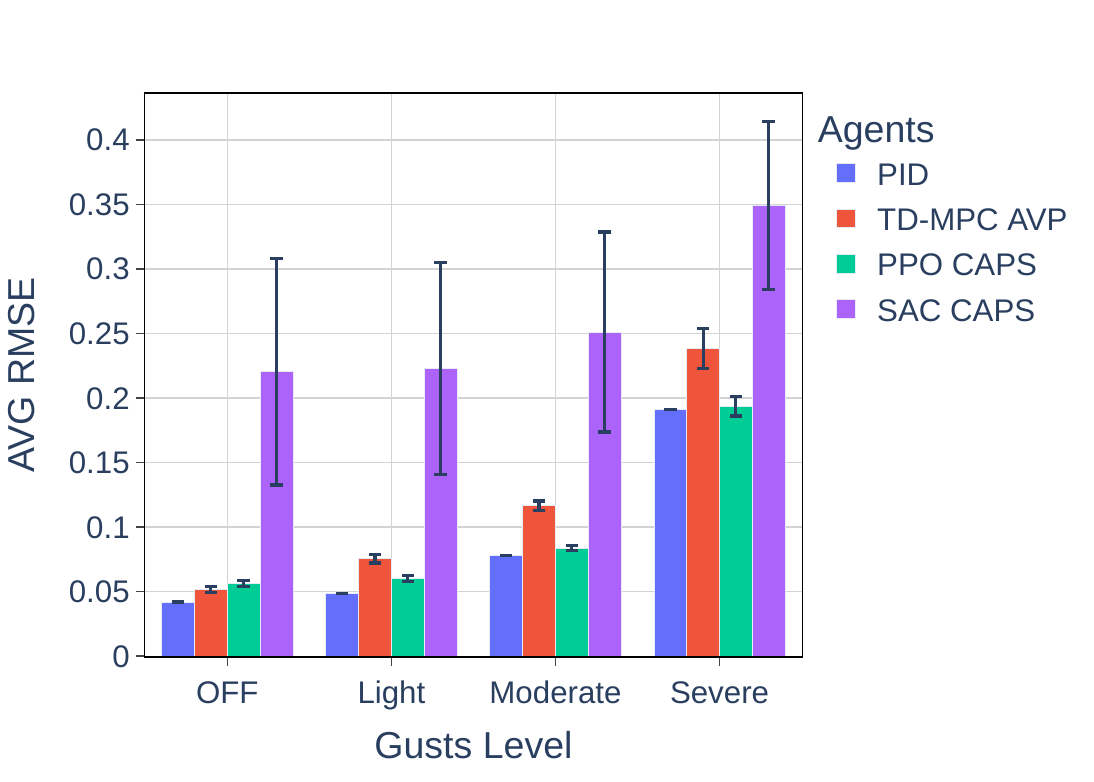}
    \caption{\textbf{Wind Gusts + Action Regulation.} Average tracking RMSE on nominal reference difficulty.}
    \label{fig:res-gusts-barplot}
\end{figure}

The results presented in Figure \ref{fig:res-gusts-barplot} point that SAC significantly underperforms relative to the other agents and TD-MPC slightly underperforms PPO and PID. To explain these observations, we emit the hypothesis that such a disturbance alters the MDP formalism into a Partially Observable Markov Decision Process (PO-MDP). In fact, an MDP is comprised of a transition function $P(s_{t+1} | s_t, a_t) = T(s_t, a_t)$ which does not exist at every point in time in the case of unpredictable gusts. From a formal point of view, while hard references simply lead to a more complex transition function, wind gusts actually make it impossible to rigorously define such a transition function. Wind information is hidden from the agent which cannot anticipate wind variations, thus conducing to state aliasing \cite{mccallum1996rlalias}. In fact, unpredictable gusts lead to the following situation: for two separate experiments, for a given common state, taking the same action could lead to two different next states. Recurrent Neural Networks have been presented as a good baseline for solving PO-MDPs \cite{ni2022rnn-pomdps,asri2019rnn} and could be applied to all the RL methods presented in the present article.

RL methods learn the transition function either explicitly in the case of model-based RL (TD-MPC) or implicitly in the case of model-free RL (PPO and SAC). Due to state aliasing, opposite updates of the function approximators occur at the onset of a wind gust. To explain PPO's better performance compared to its RL counterparts (SAC and TD-MPC), a second hypothesis could be PPO's clipping of the policy loss \cite{schulmanPPO}. Motivated by limiting too large policy updates that could lead to a bad policy where it can be hard to recover from, this conservative policy update could protect the PPO agent from the contradictory updates when subject to wind gusts.

\subsection{Discussions}
\label{res:discussions}
\textbf{TD-MPC's superiority for nonlinear regimes.} The superior results of TD-MPC under nonlinear dynamics are consistent with the reported superior performance of TD-MPC over other MF-RL methods for complex nonlinear manipulation and locomotion tasks \cite{Hansen2022tdmpc,hansen2024tdmpc2}. It solidifies the hypothesis that mixed control, combining learning an explicit model of the system dynamics and leveraging it for planning with terminal value estimation, can result in improved tracking performance for a wide range of attitude angles. It also suggests that learning an implicit representation of the dynamics limits the ability of MF-RL agents to deal with different dynamic regimes when trained simultaneously on nominal and hard regimes.

\textbf{Wind Disturbances.} We also evaluated the control methods under wind disturbances of different nature. Both studies revealed that with the exception of SAC underperforming, the severity of turbulences dominates over the choice of algorithm to predict final performances. As previously stated, hard references only consist in a harder-to-model nonlinear MDP transition function. We hypothesized that turbulence and gusts transform the MDP formalism into a non-stationary PO-MDP and we observe that not all perturbations affect with equal magnitude how well the MDP formalization fits the environment conditions. The turbulence study presented in Section \ref{subsubsec:res-stochastic-turb} suggest that turbulence can be dealt with more simply than gusts because there exists a time-averaged MDP close to the true MDP that RL algorithms fall back into by default. However, gusts present a multi-modal problem (nominal mode and gust mode) by making it impossible to strictly define a transition function which could be detrimental in the learning phase as presented in Section \ref{subsubsec:res-wind-gusts}. Therefore, we conclude that training agents with various of perturbations is not enough to ensure robustness and generalization and that different types of perturbations may require different types of approaches. The multi-model MDP framework could be investigated for this purpose \cite{steimle2021multi}.

\section{CONCLUSIONS AND PERSPECTIVES}
\label{sec:conclusion}
We proposed an in-depth study of different RL controllers with the aim of comparing model-free and model-based RL approaches. {TD-MPC}, an MB-RL method with algorithmic elements from CT and RL, yielded the best performance in nominal wind conditions. Its superiority especially shined for hard references and demonstrated its ability to perform well for deterministic nonlinear dynamics across the entire state-space. We attribute such results to TD-MPC's learning of an explicit dynamics model jointly used with predictive planning. We evaluated these control methods under various wind perturbations and found that, aside from SAC underperforming, turbulence severity significantly impacts final performance more than the choice of algorithm.

We identified high actuation fluctuation as an important drawback of RL methods. Since this metric is is key in robotics, we applied and tested countermeasures, namely: action variation reward penalty (AVP) for all methods and regularization of the actor network through CAPS for the MF-RL methods. As a result, we retained CAPS as action regulation method due to its consistency and ease of tuning across all MF-RL agents. Another path to smooth actions for TD-MPC (for which CAPS is not applicable) is MoDeM-v2 \cite{lancaster2023modem} which aims at learning safe policies by biasing the initial data distribution towards a desired behavior, in our case an action smooth controller.

 % We hypothesize that turbulence and gusts transform the MDP into a non-stationary PO-MDP. Turbulence can be managed more easily than gusts, as it allows RL algorithms to revert to a time-averaged MDP. In contrast, gusts create a multi-modal problem, complicating the learning phase. Thus, training with various perturbations alone is insufficient for robustness and different perturbations may require distinct approaches.

We identify several future research directions: one could focus on experimenting with probabilistic models for TD-MPC in order to better capture the stochasticity of turbulence dynamics, thus formulating a more rigorous transition function that outputs a probability distribution. For both gusts and turbulence, RL algorithms with recurrent networks appear to be a good starting point \cite{ni2022rnn-pomdps,asri2019rnn} to solve PO-MDPs. One could also use ideas from robust-RL to achieve disturbance resiliency of RL controllers \cite{hsu2024reforma}. Other directions include learning-based adaptive control as a field of potential mixed control methods \cite{shi2019neural,doukhi2019neural}, where a closed form of the nominal dynamics is used together with a feed-forward component of the unknown, disturbance dynamics predicted by a learned model.

\bibliographystyle{apalike}
{\small
\bibliography{example}}

\begin{thebibliography}{}

\bibitem[mil, 1980]{milspec}
 (1980).
\newblock Flying qualities of piloted airplanes, military specification.
\newblock Technical report, MIL-F-8785C.

\bibitem[Andrychowicz et~al., 2020]{andrychowicz2020learning}
Andrychowicz, O.~M., Baker, B., Chociej, M., Jozefowicz, R., McGrew, B., Pachocki, J., Petron, A., Plappert, M., Powell, G., Ray, A., et~al. (2020).
\newblock Learning dexterous in-hand manipulation.
\newblock {\em The International Journal of Robotics Research}, 39(1):3--20.

\bibitem[Asri and Trischler, 2019]{asri2019rnn}
Asri, L.~E. and Trischler, A. (2019).
\newblock A study of state aliasing in structured prediction with {RNNs}.
\newblock {\em arXiv preprint arXiv:1906.09310}.

\bibitem[Beard and McLain, 2012]{beard2012small}
Beard, R.~W. and McLain, T.~W. (2012).
\newblock {\em Small Unmanned Aircraft: Theory and Practice}.
\newblock Princeton University Press.

\bibitem[Becker-Ehmck et~al., 2020]{becker2020learning}
Becker-Ehmck, P., Karl, M., Peters, J., and van~der Smagt, P. (2020).
\newblock Learning to fly via deep model-based reinforcement learning.
\newblock {\em arXiv preprint arXiv:2003.08876}.

\bibitem[Berndt, 2004]{berndt2004jsbsim}
Berndt, J. (2004).
\newblock {JSBSim}: An open source flight dynamics model in {C++}.
\newblock In {\em AIAA Modeling and Simulation Technologies Conference and Exhibit}, page 4923.

\bibitem[B{\o}hn et~al., 2019]{bohn2019deep}
B{\o}hn, E., Coates, E.~M., Moe, S., and Johansen, T.~A. (2019).
\newblock Deep reinforcement learning attitude control of fixed-wing {UAV}s using proximal policy optimization.
\newblock In {\em International Conference on Unmanned Aircraft Systems (ICUAS)}, pages 523--533.

\bibitem[B{\o}hn et~al., 2023]{bohn2023data}
B{\o}hn, E., Coates, E.~M., Reinhardt, D., and Johansen, T.~A. (2023).
\newblock Data-efficient deep reinforcement learning for attitude control of fixed-wing {UAV}s: Field experiments.
\newblock {\em IEEE Transactions on Neural Networks and Learning Systems}.

\bibitem[De~Marco et~al., 2023]{de2023deep}
De~Marco, A., D’Onza, P.~M., and Manfredi, S. (2023).
\newblock A deep reinforcement learning control approach for high-performance aircraft.
\newblock {\em Nonlinear Dynamics}, 111(18):17037--17077.

\bibitem[Doukhi and Lee, 2019]{doukhi2019neural}
Doukhi, O. and Lee, D.~J. (2019).
\newblock Neural network-based robust adaptive certainty equivalent controller for quadrotor {UAV} with unknown disturbances.
\newblock {\em International Journal of Control, Automation and Systems}, 17(9):2365--2374.

\bibitem[Gryte et~al., 2018]{gryte2018aerodynamic}
Gryte, K., Hann, R., Alam, M., Roh{\'a}{\v{c}}, J., Johansen, T.~A., and Fossen, T.~I. (2018).
\newblock Aerodynamic modeling of the {Skywalker x8} fixed-wing unmanned aerial vehicle.
\newblock In {\em International Conference on Unmanned Aircraft Systems (ICUAS)}, pages 826--835.

\bibitem[Haarnoja et~al., 2018]{haarnoja2018sac}
Haarnoja, T., Zhou, A., Abbeel, P., and Levine, S. (2018).
\newblock Soft actor-critic: Off-policy maximum entropy deep reinforcement learning with a stochastic actor.
\newblock In {\em International Conference on Machine Learning}, pages 1861--1870. PMLR.

\bibitem[Hafner et~al., 2023]{hafner2023dreamerv3}
Hafner, D., Pasukonis, J., Ba, J., and Lillicrap, T. (2023).
\newblock Mastering diverse domains through world models.
\newblock {\em arXiv preprint arXiv:2301.04104}.

\bibitem[Hansen et~al., 2024]{hansen2024tdmpc2}
Hansen, N., Su, H., and Wang, X. (2024).
\newblock {TD-MPC2}: Scalable, robust world models for continuous control.
\newblock In {\em International Conference on Learning Representations (ICLR)}.

\bibitem[Hansen et~al., 2022]{Hansen2022tdmpc}
Hansen, N., Wang, X., and Su, H. (2022).
\newblock Temporal difference learning for model predictive control.
\newblock In {\em International Conference on Machine Learning}.

\bibitem[Hsu et~al., 2024]{hsu2024reforma}
Hsu, H.-L., Meng, H., Luo, S., Dong, J., Tarokh, V., and Pajic, M. (2024).
\newblock {REFORMA}: Robust reinforcement learning via adaptive adversary for drones flying under disturbances.
\newblock In {\em 2024 IEEE International Conference on Robotics and Automation (ICRA)}.

\bibitem[Huang et~al., 2022]{huang2022cleanrl}
Huang, S., Dossa, R. F.~J., Ye, C., Braga, J., Chakraborty, D., Mehta, K., and Araújo, J.~G. (2022).
\newblock {CleanRL}: High-quality single-file implementations of deep reinforcement learning algorithms.
\newblock {\em Journal of Machine Learning Research}, 23(274):1--18.

\bibitem[Hwangbo et~al., 2017]{hwangbo2017control}
Hwangbo, J., Sa, I., Siegwart, R., and Hutter, M. (2017).
\newblock Control of a quadrotor with reinforcement learning.
\newblock {\em IEEE Robotics and Automation Letters}, 2(4):2096--2103.

\bibitem[Koch et~al., 2019]{koch2019reinforcement}
Koch, W., Mancuso, R., West, R., and Bestavros, A. (2019).
\newblock Reinforcement learning for {UAV} attitude control.
\newblock {\em ACM Transactions on Cyber-Physical Systems}, 3(2):1--21.

\bibitem[Lambert et~al., 2019]{lambert2019low}
Lambert, N.~O., Drew, D.~S., Yaconelli, J., Levine, S., Calandra, R., and Pister, K.~S. (2019).
\newblock Low-level control of a quadrotor with deep model-based reinforcement learning.
\newblock {\em IEEE Robotics and Automation Letters}, 4(4):4224--4230.

\bibitem[Lancaster et~al., 2023]{lancaster2023modem}
Lancaster, P., Hansen, N., Rajeswaran, A., and Kumar, V. (2023).
\newblock {MoDem-V2}: Visuo-motor world models for real-world robot manipulation.
\newblock {\em arXiv preprint arXiv:2309.14236}.

\bibitem[Lee et~al., 2020]{lee2020learning}
Lee, J., Hwangbo, J., Wellhausen, L., Koltun, V., and Hutter, M. (2020).
\newblock Learning quadrupedal locomotion over challenging terrain.
\newblock {\em Science robotics}, 5(47).

\bibitem[Liang et~al., 2018]{liang2018scalablembrl}
Liang, X., Zheng, M., and Zhang, F. (2018).
\newblock A scalable model-based learning algorithm with application to uavs.
\newblock {\em IEEE control systems letters}, 2(4):839--844.

\bibitem[Liu et~al., 2021]{liu2021learning}
Liu, F., Dai, S., and Zhao, Y. (2021).
\newblock Learning to have a civil aircraft take off under crosswind conditions by reinforcement learning with multimodal data and preprocessing data.
\newblock {\em Sensors}, 21(4):1386.

\bibitem[Mathisen et~al., 2021]{mathisen2021precision}
Mathisen, S., Gryte, K., Gros, S., and Johansen, T.~A. (2021).
\newblock Precision deep-stall landing of fixed-wing uavs using nonlinear model predictive control.
\newblock {\em Journal of Intelligent \& Robotic Systems}, 101:1--15.

\bibitem[McCallum, 1996]{mccallum1996rlalias}
McCallum, A.~K. (1996).
\newblock {\em Reinforcement learning with selective perception and hidden state}.
\newblock University of Rochester.

\bibitem[Mnih et~al., 2013]{mnih2013atari}
Mnih, V., Kavukcuoglu, K., Silver, D., Graves, A., Antonoglou, I., Wierstra, D., and Riedmiller, M. (2013).
\newblock Playing {Atari} with deep reinforcement learning.
\newblock {\em arXiv preprint arXiv:1312.5602}.

\bibitem[Moallemi and Towhidnejad, 2016]{moallemi2016b}
Moallemi, M. and Towhidnejad, M. (2016).
\newblock B-737 autopilot design and implementation for simulated flight management system.
\newblock In {\em Proceedings of the 49th Annual Simulation Symposium}, pages 1--7.

\bibitem[Mysore et~al., 2021]{mysore2021regularizing}
Mysore, S., Mabsout, B., Mancuso, R., and Saenko, K. (2021).
\newblock Regularizing action policies for smooth control with reinforcement learning.
\newblock In {\em IEEE International Conference on Robotics and Automation (ICRA)}, pages 1810--1816.

\bibitem[Ni et~al., 2022]{ni2022rnn-pomdps}
Ni, T., Eysenbach, B., and Salakhutdinov, R. (2022).
\newblock Recurrent model-free {RL} can be a strong baseline for many pomdps.
\newblock In {\em International Conference on Machine Learning}, pages 16691--16723. PMLR.

\bibitem[Peng et~al., 2020]{peng2020learning}
Peng, X.~B., Coumans, E., Zhang, T., Lee, T.-W., Tan, J., and Levine, S. (2020).
\newblock Learning agile robotic locomotion skills by imitating animals.
\newblock {\em arXiv preprint arXiv:2004.00784}.

\bibitem[Schrittwieser et~al., 2020]{schrittwieser2020mastering}
Schrittwieser, J., Antonoglou, I., Hubert, T., Simonyan, K., Sifre, L., Schmitt, S., Guez, A., Lockhart, E., Hassabis, D., Graepel, T., et~al. (2020).
\newblock Mastering {Atari, Go, Chess and Shogi} by planning with a learned model.
\newblock {\em Nature}, 588(7839):604--609.

\bibitem[Schulman et~al., 2017]{schulmanPPO}
Schulman, J., Wolski, F., Dhariwal, P., Radford, A., and Klimov, O. (2017).
\newblock Proximal policy optimization algorithms.
\newblock {\em arXiv preprint arXiv:1707.06347}.

\bibitem[Seyde et~al., 2021]{seyde2021bang}
Seyde, T., Gilitschenski, I., Schwarting, W., Stellato, B., Riedmiller, M., Wulfmeier, M., and Rus, D. (2021).
\newblock Is bang-bang control all you need? solving continuous control with bernoulli policies.
\newblock {\em Advances in Neural Information Processing Systems}, 34:27209--27221.

\bibitem[Shi et~al., 2021]{shi2021omac}
Shi, G., Azizzadenesheli, K., O'Connell, M., Chung, S.-J., and Yue, Y. (2021).
\newblock Meta-adaptive nonlinear control: Theory and algorithms.
\newblock {\em Advances in Neural Information Processing Systems}, 34:10013--10025.

\bibitem[Shi et~al., 2019]{shi2019neural}
Shi, G., Shi, X., O’Connell, M., Yu, R., Azizzadenesheli, K., Anandkumar, A., Yue, Y., and Chung, S.-J. (2019).
\newblock Neural lander: Stable drone landing control using learned dynamics.
\newblock In {\em International Conference on Robotics and Automation}, pages 9784--9790.

\bibitem[Song et~al., 2023a]{song2023lipsnet}
Song, X., Duan, J., Wang, W., Li, S.~E., Chen, C., Cheng, B., Zhang, B., Wei, J., and Wang, X.~S. (2023a).
\newblock {LipsNet}: a smooth and robust neural network with adaptive {L}ipschitz constant for high accuracy optimal control.
\newblock In {\em International Conference on Machine Learning}, pages 32253--32272. PMLR.

\bibitem[Song et~al., 2023b]{song2023reaching}
Song, Y., Romero, A., M{\"u}ller, M., Koltun, V., and Scaramuzza, D. (2023b).
\newblock Reaching the limit in autonomous racing: Optimal control versus reinforcement learning.
\newblock {\em Science Robotics}, 8(82).

\bibitem[Steimle et~al., 2021]{steimle2021multi}
Steimle, L.~N., Kaufman, D.~L., and Denton, B.~T. (2021).
\newblock Multi-model {M}arkov decision processes.
\newblock {\em IISE Transactions}, 53(10):1124--1139.

\bibitem[Sutton and Barto, 2018]{sutton2018reinforcement}
Sutton, R.~S. and Barto, A.~G. (2018).
\newblock {\em Reinforcement learning: An introduction}.
\newblock MIT press.

\bibitem[Tsourdos et~al., 2019]{tsourdos2019developing}
Tsourdos, A., Permana, I. A.~D., Budiarti, D.~H., Shin, H.-S., and Lee, C.-H. (2019).
\newblock Developing flight control policy using deep deterministic policy gradient.
\newblock In {\em IEEE International Conference on Aerospace Electronics and Remote Sensing Technology (ICARES)}.

\bibitem[Williams et~al., 2015]{williams2015MPPI}
Williams, G., Aldrich, A., and Theodorou, E. (2015).
\newblock Model predictive path integral control using covariance variable importance sampling.
\newblock {\em arXiv preprint arXiv:1509.01149}.

\end{thebibliography}

%\section*{\uppercase{Appendix}}

\end{document}